\documentclass{article}

\usepackage{microtype}
\usepackage{graphicx}
\usepackage{subfigure}
\usepackage{booktabs} 
\usepackage{ulem} 
\usepackage{multirow}
\usepackage{enumitem}
\usepackage[ruled,vlined,linesnumbered]{algorithm2e}

\usepackage{hyperref}

\usepackage[accepted]{icml2025}

\usepackage{amsmath}
\usepackage{amssymb}
\usepackage{mathtools}
\usepackage{amsthm}

\usepackage[capitalize,noabbrev]{cleveref}

\theoremstyle{plain}

\theoremstyle{definition}

\theoremstyle{remark}

\SetAlgoLined
\SetAlgoNlRelativeSize{0}
\SetNlSty{textbf}{\color{black}}{}
\SetAlFnt{\small}
\SetAlCapFnt{\small}
\SetAlCapNameFnt{\small}
\SetAlCapHSkip{0pt}
\IncMargin{-\parindent}
\SetKwComment{Comment}{$\triangleright$ }{}
\SetCommentSty{color{commentcolor}}
\SetKwInOut{Input}{\textbf{Input}} 
\SetKwInOut{Output}{\textbf{Output}} 
\SetKw{KwTo}{in}
\SetKwFor{ForEach}{for each}{do}{end}
\SetKwIF{If}{ElseIf}{Else}{if}{then}{else if}{else}{end}
\SetKwFunction{FMain}{FindSimilar}
\SetKwProg{Fn}{Function}{:}{end}

\usepackage[textsize=tiny]{todonotes}

\icmltitlerunning{Submission and Formatting Instructions for ICML 2025}

\begin{document}

\twocolumn[
\icmltitle{Multivariate de Bruijn Graphs: A Symbolic Graph Framework for\\ Time Series Forecasting}



\icmlsetsymbol{equal}{}

\begin{icmlauthorlist}
\icmlauthor{Mert Onur Cakiroglu}{yyy}
\icmlauthor{Idil Bilge Altun}{yyy}
\icmlauthor{Mehmet Dalkilic}{yyy}
\icmlauthor{Elham Buxton}{sch}
\icmlauthor{Hasan Kurban}{comp}

\end{icmlauthorlist}

\icmlaffiliation{yyy}{Department of Computer Science, Indiana University Bloomington, IN, USA}
\icmlaffiliation{comp}{College of Science and Engineering, Hamad Bin Khalifa University, Doha, Qatar}
\icmlaffiliation{sch}{Department of Computer Science, University of Illinois Springfield, IL, USA}

\icmlcorrespondingauthor{Hasan Kurban}{hkurban@hbku.edu.qa}

\icmlkeywords{Machine Learning, ICML}

\vskip 0.3in
]



\printAffiliationsAndNotice{} 

\begin{abstract}
Time series forecasting remains a challenging task for foundation models due to temporal heterogeneity, high dimensionality, and the lack of inherent symbolic structure. In this work, we propose DRAGON (\uline{D}iscrete \uline{R}epresentation and \uline{A}ugmented \uline{G}raph encoding \uline{O}ver de Bruij\uline{N} Graphs), a novel encoder that introduces Multivariate de Bruijn Graphs (MdBGs) to bridge the gap between symbolic representations and neural modeling. DRAGON discretizes continuous input sequences and maps them onto a fixed graph structure, enabling dynamic context recovery via graph-based attention. Integrated as an auxiliary module within a dual-branch architecture, DRAGON augments conventional CNN-based encoders with symbolic, structure-aware representations. All code developed for this study is available at: {\url{https://github.com/KurbanIntelligenceLab/MultdBG-Time-Series-Library}}
\end{abstract}

\section{Introduction}
Foundation models have made great strides in NLP \cite{zhao2023survey} and computer vision \cite{awais2023foundational} by utilizing non-numeric, large-scale data; however, the same level of generalization has not emerged in the domain of time series where challenges exist \textit{e.g.}, temporal heterogeneity, multivariate dependencies, and distant relationships.

\begin{figure}[!ht]
    \centering
    \includegraphics[width=1\linewidth]{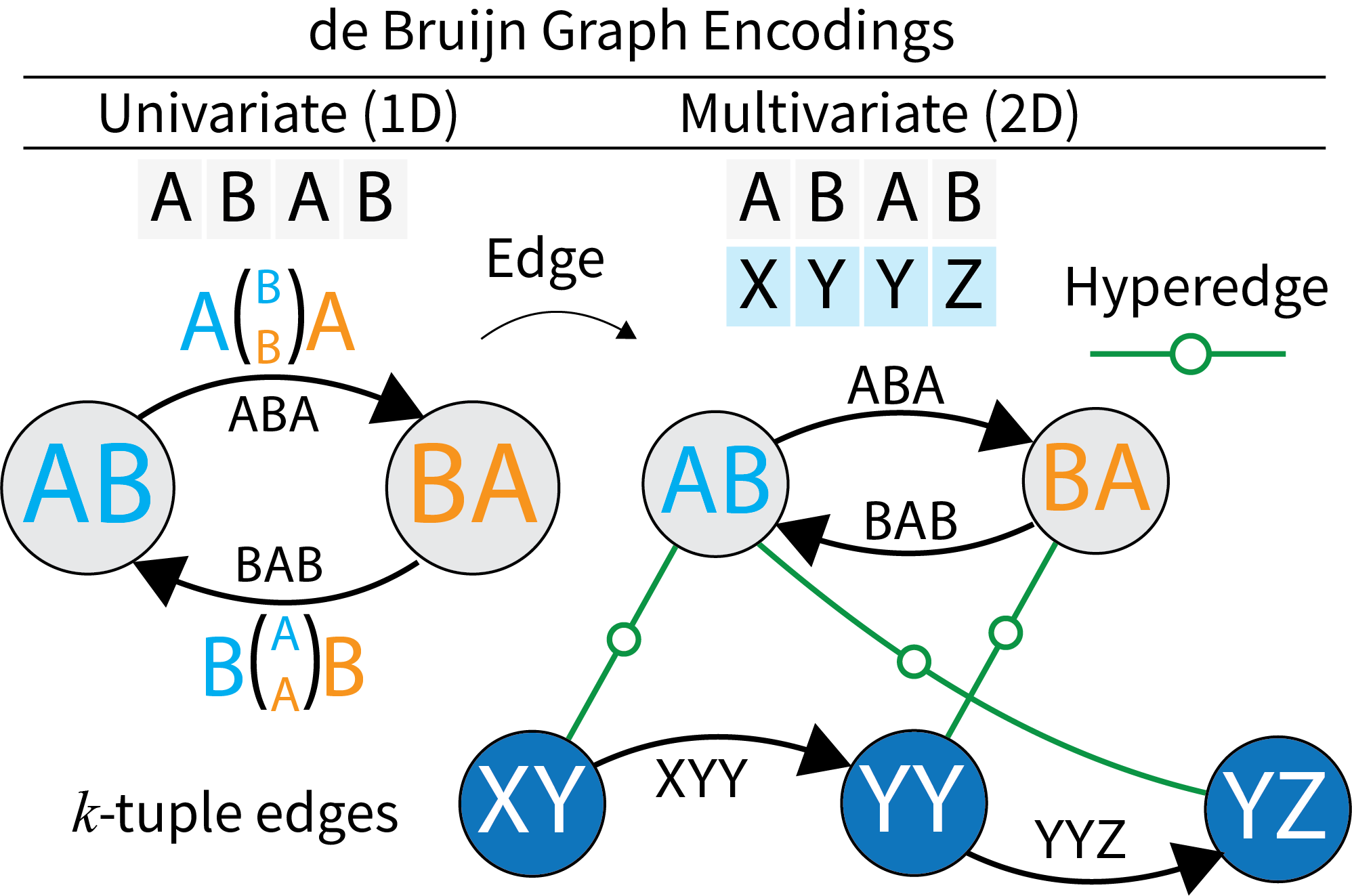}
    \vspace{-.8cm}
    \caption{Illustration of de Bruijn Graphs for univariate and multivariate dimensions. ($\textsc{Left}$) The univariate dBG encodes $k$-tuples with directed and weighted edges in a single dimension. ($\textsc{Right}$) The multivariate dBG extends this structure by incorporating edges (hyper-tuples) that connect $k$-tuples across multiple dimensions at the same time step, capturing inter-variable dependencies.}
    \label{fig:MdBG}
\end{figure}

Existing time series forecasting models rely heavily on the inputs being numeric, especially continuous \cite{ekambaram2024tsmixer} \cite{nie2023patchtst} while missing out on recurring motifs that are central to many real-world temporal processes. Moreover, the common practice of creating fixed-size sliding window fragments from long sequences to facilitate supervised training limits the model's access to the global context outside the current fragment. To address these gaps, we introduce a new architecture that incorporates a symbolic graph-based structure into time series modeling through de Bruijn Graphs (dBG) \cite{de1946combinatorial}. dBGs are widely used in computational biology in applications such as genome assembly and k-mer–based (also known as k-tuple) sequence modeling to efficiently and compactly represent long sequences from overlapping fragments \cite{compeau2011bruijn}. Recently, dBGs have also been explored in the context of time series modeling for capturing symbolic temporal patterns and enhancing forecasting performance \cite{10722826,Cakiroglu2024}. Modeling time series data as dBG allows transforming continuous temporal data into a compact symbolic representation that captures recurring motifs and long-range dependencies across sequence fragments. We introduce a novel method for encoding multivariate time series as Multivariate de Bruijn Graphs (MdBGs) to capture patterns both within and across dimensions, allowing models to access cross sequence global context while processing each sliding window locally.

Our contributions are threefold: (i) MdBG, a multivariate extension of de Bruijn graphs, captures intra-, inter-dimensional, and long-range dependencies; (ii) DRAGON, an encoder built on MdBGs, generates task-specific time series embeddings and integrates into any model; and (iii) DRAGON outperforms state-of-the-art methods on forecasting benchmarks, including TimesNet \cite{wu2022timesnet}.
%
%
%
   

\label{submission}

\section{Methods}
\subsection{Multivariate de Bruijn Graphs (MdBGs)}
\subsubsection{Data Processing}

 Let $\mathcal{D}^{\text{raw}} = \{\mathbf{X}_1^{\text{raw}}, \mathbf{X}_2^{\text{raw}}, \dots, \mathbf{X}_D^{\text{raw}}\}$ be a time series data with $D$ dimensions. Each raw sequence is defined as: $\mathbf{X}_i^{\text{raw}} = [x_1^{i}, x_2^{i}, \dots, x_S^{i}], $
where $x_t^{i} \in \mathbb{R}$ and $S$ is the sequence length. Since dBGs operate on categorical data, the raw input $\mathbf{X}_i^{\text{raw}}$ is first discretized using a discretization function $\textsc{DISC}_i(\mathbf{X}_i^{\text{raw}}, \alpha_i) \rightarrow \mathbf{X}_i^{\text{disc}}$. The set of bin sizes for all dimensions is denoted by $A = \{ \alpha_1, \alpha_2, \dots, \alpha_n \}$, where $\alpha_i$ represents the bin size for dimension $i$. Similarly, the set of discretization functions is denoted by $\mathbf{F} = \{ \textsc{DISC}_1, \textsc{DISC}_2, \dots, \textsc{DISC}_D \}$. This formulation allows discretization to be customized independently for each dimension. The resulting discretized dataset, denoted as $\mathcal{D}$, consists of sequences defined as follows: $ \mathcal{D}^{\text{disc}} = \{\mathbf{X}_1^{\text{disc}}, \mathbf{X}_2^{\text{disc}}, \dots, \mathbf{X}_D^{\text{disc}}\}, $ where $\mathbf{X}_i^{\text{disc}} = [c_1^{i}, c_2^{i}, \dots, c_S^{i}], \quad c_t^{i} \in \{1, 2, \dots, \alpha_i\}.$ Each discretized element $c_t^{i}$ represents a categorical value bounded by the discretization level $\alpha_i$.

\begin{figure}[!ht]
    \centering
    \includegraphics[width=1\linewidth]{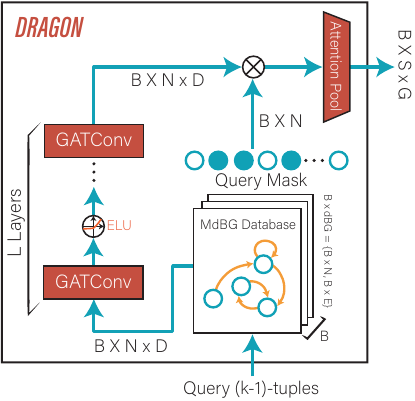}
    \caption{The DRAGON architecture employs a MdBG as a fixed graph database built from the entire training set. MdBG structure remains constant across batches and is encoded through $L$ layers of Graph Attention Convolution (GATConv) with ELU activation. A node-level hard mask determines active nodes and is used to select encodings for the subgraph corresponding to the$(k{-}1)$-tuples in the input sequence. Only the node features are updated through masking. To obtain the final output, an attention pooling mechanism is applied that aggregates the graph-level features and reshapes them to match the input sequence length. $B$ denotes the Batch size, $D$ is the number of input dimensions, $N$ is the number of MdBG nodes and $E$ is the edge feature dimension.}
    \label{fig:DRAGON}
\end{figure}

\begin{figure*}[!ht]
    \centering
    \includegraphics[width=1\linewidth]{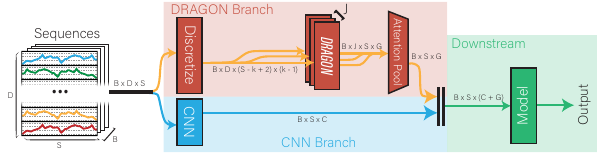}
    \caption{Overview of the proposed DRAGON architecture for multivariate time series modeling. The input sequence of shape $\mathbb{R}^{B \times D \times S}$ where B is the batch size, is processed through two parallel branches. In the top branch, the sequence is discretized and passed into the DRAGON encoder, which constructs MdBGs and generates graph-based embeddings of shape $\mathbb{R}^{B \times J \times S \times G}$ where $J$ is the number of the DRAGON encoders. These embeddings are then aggregated via an attention pooling mechanism to produce the output of the shape $\mathbb{R}^{B \times S \times G}$. In the bottom branch, the original continuous sequence is fed into a 1D CNN encoder, yielding features of shape $\mathbb{R}^{B \times S \times C}$. The outputs from both branches are concatenated to form a joined representation of shape $\mathbb{R}^{B \times S \times (C + G)}$, which is then sent to a downstream model for prediction.}
    \label{fig:Downstream}
\end{figure*}

\subsubsection{Graph Construction}

A single MdBG is constructed from all sequences in the training set prior to model training. This allows the entire training set to be compactly represented as a graph and provides global context during model training. MdBG is a structured graph representation of both intra- and inter-dimensional dependencies of multivariate sequence data. It is formally defined as  \(G(V, E)\) where $V$ is the set of nodes (fixed-length strings over some alphabet)  and $E$ is the set of directed edges (fixed-length strings over the same alphabet) where some  overlapping substring of fixed size between two nodes. $E$ is labeled with a non-negative number.   

More formally, a dBG of order $k$ is a directed graph where each node represents a unique univariate $(k{-}1)$-tuple (a subsequence of length $k{-}1$), and each edge corresponds to a $k$-tuple (length-$k$ subsequence) observed in the input sequence. Specifically, the set of $k$-tuples $\mathbf{e}_t^{i} = \left(x_t^{i}, x_{t+1}^{i}, \dots, x_{t+k-1}^{i}\right)$ induces a directed edge from the prefix node to the suffix node, $\mathbf{v}_{\text{prefix}} = (x_t^{i}, \dots, x_{t+k-2}^{i}) \rightarrow \mathbf{v}_{\text{suffix}} = (x_{t+1}^{i}, \dots, x_{t+k-1}^{i})$ for $t = 1, 2, \dots, \left( S - k + 1 \right)$. Each $\mathbf{e}_t^{i}$ defines a directed edge between two $(k{-}1)$-tuple nodes, capturing temporal transitions in the sequence. Repeated transitions increment the edge weights, encoding frequency. The construction of a dBG from univariate data is illustrated in Figure \ref{fig:MdBG} ($\textsc{Left}$).

Each node $\mathbf{v} \in V$ can correspond to a set of continuous $k$-tuples extracted from the original raw dataset $\mathcal{D}^{\text{raw}}$. This set is denoted as the feature space of node $\mathbf{v}$:

$$
\mathcal{F}_{\mathbf{v}} = \{x^k_1, x^k_2, \dots, x^k_m\},
$$

where each $x^k_i \in \mathbb{R}^k$ and $1 \leq m \leq S - k + 2$. This feature set $\mathcal{F}_{\mathbf{v}}$ captures the continuous embeddings associated with the discrete representation $\mathbf{v}$.

After constructing individual dBGs for each dimension, a set of $D$ disconnected subgraphs $\textbf{MdBG} = \{ G_1, G_2, \dots, G_D \}$ is obtained, where each graph layer $G_i = (V_i, E_i)$ contains a set of nodes $V_i$ representing all observed univariate $(k{-}1)$-tuples from dimension $i$, and directed edges $E_i$ corresponding to observed $k$-tuples.

To capture inter-dimensional structure, hyper-tuple edges that connect nodes across different dimensions are defined. Given two nodes $\mathbf{v} \in V_i$ and $\mathbf{u} \in \mathcal{V}_j$, a bi-directional edge is established if their feature sets $\mathcal{F}_{\mathbf{v}_i}$ and $\mathcal{F}_{\mathbf{u}_j}$ contain raw $k$-tuples that occur at the same time step. This construction enables the graph to represent temporal dependencies across dimensions. As a result, for each multi-dimensional point $x \in \mathbb{R}^D$, a hyper-tuple clique of size $D$ is formed by linking all co-occurring $(k{-}1)$-tuples across the $D$ dimensions. This process is illustrated in Figure \ref{fig:MdBG} ($\textsc{Right}$). Note that the definition of MdBG is an extension of the original dBG; when $D = 1$, the MdBG is structurally equivalent to the standard dBG structure. An algorithm for constructing the MdBG is presented in Algorithm~\ref{alg:mdBG-construction} in the Appendix.

Once MdBG is constructed from the training set, Graph Diffusion Convolution (GDC) is applied to further refine the structural properties of the MdBG and enhance information propagation across the constructed multivariate graph. GDC modifies the adjacency matrix by incorporating global diffusion patterns, effectively capturing long-range dependencies and smoothing over noisy or sparse connections. Specifically, we utilize Personalized PageRank (PPR)-based diffusion, which has demonstrated improved performance in node representation learning tasks by preserving both local and global graph structure \cite{GDC}. This transformation improves the robustness and expressiveness of the learned node embeddings, which are later used in downstream tasks.

\begin{table*}[!ht]
\centering
\caption{Average MSE and MAE per model across four datasets (ETTh1, ETTh2, ETTm1, ETTm2). Lowest values in each column are highlighted in bold. The DRAGON module achieves the lowest error rates in the majority of settings, demonstrating its effectiveness in recovering long-range dependencies from severely limited input contexts. These results are obtained using a fixed input length of 12 and illustrate the generalization capabilities of DRAGON in multivariate time series benchmarks.}
\label{tab:summary_results}
        \scalebox{0.9}{
\begin{tabular}{lcc | cc | cc | cc}
\toprule
 & \multicolumn{2}{c|}{\textbf{ETTh1}} & \multicolumn{2}{c|}{\textbf{ETTh2}} & \multicolumn{2}{c|}{\textbf{ETTm1}} & \multicolumn{2}{c}{\textbf{ETTm2}} \\
\textbf{Model} & \textbf{MSE} & \textbf{MAE} & \textbf{MSE} & \textbf{MAE} & \textbf{MSE} & \textbf{MAE} & \textbf{MSE} & \textbf{MAE} \\
\midrule
Transformer \cite{transformer} & 0.909 & 0.733 & 2.54 & 1.219 & 0.889 & 0.672 & 1.404 & 0.844 \\
TSMixer  \cite{timeXer} & 1.002 & 0.756 & 2.262 & 1.236 & 0.916 & 0.690 & 0.907 & 0.723 \\
Crossformer \cite{ekambaram2024tsmixer} & 0.691 & 0.609 & 1.982 & 1.060 & \textbf{0.718} & 0.588 & 1.005 & 0.675 \\
FiLM \cite{film} & 0.712 & 0.550 & 0.450 & 0.433 & 1.031 & 0.626 & 0.350 & 0.375 \\
Nonstationary \cite{nonstationary} & 0.742 & 0.586 & 0.576 & 0.506 & 0.764 & \uline{0.557} & 0.418 & 0.399 \\
PatchTST \cite{nie2023patchtst} & 0.642 & 0.525 & 0.448 & 0.432 & 1.014 & 0.624 & 0.347 & 0.373 \\
TimeMixer \cite{timemixer} & 0.633 & 0.519 & 0.447 & 0.434 & 0.942 & 0.601 & 0.343 & 0.370 \\
Autoformer \cite{wu2021autoformer} & 0.656 & 0.556 & 0.458 & 0.450 & 0.854 & 0.603 & 0.335 & 0.376 \\
TimeXer \cite{timeXer} & \uline{0.613} & \uline{0.512} & \textbf{0.444} & \textbf{0.428} & 0.904 & 0.589 & 0.336 & 0.362 \\
TimesNet \cite{wu2022timesnet} & 0.639 & 0.525 & 0.457 & 0.438 & 0.850 & 0.578 & \uline{0.332} & \uline{0.358} \\
DRAGON (Ours) & \textbf{0.604} & \textbf{0.510} & \uline{0.446} & \uline{0.430} & \uline{0.742} & \textbf{0.542} & \textbf{0.320} & \textbf{0.348} \\
\bottomrule
\end{tabular}}
\end{table*}

\subsection{Training}
During training, the DRAGON module encodes a subgraph of MdBG corresponding to the current input sequence. For each input sequence, query $(k{-}1)$-tuples are generated  using a sliding window mechanism and subsequently discretized using the dimension-specific functions in the set $\mathbf{F} = \{\textsc{DISC}_1, \dots, \textsc{DISC}_D\}$. These functions map continuous input sequences into categorical representations, yielding discretized tuples consistent with the MdBG vocabulary. 

During training, all generated tuples are assumed to exist within the MdBG, the encoding for the nodes that correspond to the tuples in the input sequence can be simply retrieved. However, at test time, a query tuple $\hat{\mathbf{q}}$ may not appear in the MdBG node set $\mathcal{V}$, so the subgraph corresponding to the input sequence is approximated by selecting  the most similar node based on L1 norm (Manhattan distance) from  $\hat{\mathbf{q}}$. More specifically, masking vector $\mathbf{m}^{\mathbf{v}}$ over all nodes $\mathbf{v} \in \mathcal{V}$ is defined as follows:
$$
\mathbf{m}^{\mathbf{v}} =
\begin{cases}
1 & \text{if } \mathbf{v} = \hat{\mathbf{q}} \in \mathcal{V} \\
1 & \text{if } \mathbf{v} = \displaystyle\arg\min_{u \in \mathcal{V}} \|\hat{\mathbf{q}} - u\|_1 \quad \text{and } \hat{\mathbf{q}} \notin \mathcal{V} \\
0 & \text{otherwise}
\end{cases}
$$
This masking mechanism ensures that either the exact or the closest matching node is encoded for each $(k{-}1)$-tuple in the input sequence. The overall encoder architecture is illustrated in Figure~\ref{fig:DRAGON}.

The DRAGON module can be seamlessly integrated into most time series encoders as an auxiliary component, enhancing encoding by providing global context on the structure and recurring motifs present in the training set. It operates alongside the main architecture in a dual-branch design, with both branches later concatenated to form a fused representation. The overall process is illustrated in Figure \ref{fig:Downstream}.

\textbf{(1) DRAGON Branch:} The current batch is processed through one or more DRAGON encoders. Multiple DRAGON modules can be employed to capture graph representations with different configurations (e.g., varying $k$ values or alphabet sizes $A$). The outputs from these modules are combined using an attention pooling mechanism, resulting in a unified graph-based feature representation of shape $\mathbb{R}^{B \times S \times G}$, where $G$ denotes the embedding dimension of the DRAGON branch.

\textbf{(2) Main Branch:} The original continuous input sequence is passed through a main branch, which can be any standard time series encoder, such as a Transformer or CNN. In our design, we opt for a 1D CNN due to its simplicity and computational efficiency. This  allows us to maintain a lightweight main branch,  emphasizing the  global contextual information captured by the DRAGON module.   The CNN-based main branch  yields an output of shape $\mathbb{R}^{B \times S \times C}$, where $C$ is the number of channels. 

The outputs from both branches are concatenated along the feature dimension, resulting in a fused encoding of shape $\mathbb{R}^{B \times S \times (C + G)}$. This combined representation is subsequently fed into the downstream decoder. 

\section{Results}
To evaluate the effectiveness of the DRAGON module, a series of experiments is conducted using a narrow input context of length 12—an intentionally constrained setting designed to highlight the module’s ability to recover  temporal dependencies that span beyond the immediate input window. TimesNet is used as the base downstream model, with and without the DRAGON module, to assess the impact of our approach. The performance of the model is compared across four widely-used multivariate time series forecasting benchmarks: ETTh1, ETTh2, ETTm1, and ETTm2. The results are benchmarked against several state-of-the-art (SoTA) models as well as the baseline TimesNet model. A summary of the key results is presented in Table~\ref{tab:summary_results}, and the complete results can be found in Appendix Table~\ref{tab:full_results}.

Our experiments show that the DRAGON module consistently improves performance, particularly in scenarios characterized by severely limited input context and extended forecasting horizon. By leveraging graph-based representations of historical patterns, DRAGON dynamically recovers missing temporal dependencies during inference. In addition to outperforming the TimesNet baseline, the DRAGON-enhanced variants often achieve SoTA performance when combined with TimesNet backbone, demonstrating its versatility and generalizability across forecasting architectures.

\section{Conclusion}
In this work, we introduced DRAGON, a novel encoder architecture that incorporates MdBGs into time series modeling. By discretizing continuous inputs and mapping them onto symbolic graph structures, DRAGON enables dynamic recovery of missing temporal context, especially in constrained input scenarios. Our results across four standard benchmarks show that DRAGON substantially improves performance when paired with a strong downstream model like TimesNet. The consistent gains highlight the effectiveness of symbolic graph representations in complementing neural architectures for time series analysis. DRAGON offers a new direction for time series modeling (one that bridges discrete structure and continuous representation) and sets the stage for more symbolic-aware foundation models in temporal domains.



\newpage
\appendix
\onecolumn

\section{MdBG Feature Selection}
In the Multivariate de Bruijn Graph (MdBG), nodes do not have a fixed feature set due to the discretized nature of the graph. Specifically, multiple raw $(k{-}1)$-tuples can map to the same node, resulting in feature sets of varying sizes, i.e., $1 \leq |\mathcal{F}_{\mathbf{v}}| \leq S - k + 2$. Some nodes may contain a large number of associated raw tuples in their feature set $\mathcal{F}_{\mathbf{v}}$. To handle this variability, a fixed number of $(k{-}1)$-tuples (denoted by $f$) from each node at every iteration is randomly sampled (with replacement). This strategy ensures computational tractability and encourages the model to explore diverse node features during training. $f=16$ is set for our experimentation.
 
\section{MdBG Construction Algorithm}
The MdBG construction algorithm with the time complexity of $\mathcal{O}(m^2 \cdot T)$ is shown in Algorithm \ref{alg:mdBG-construction}. The input consists of multivariate continous time series data \( \mathcal{D}^{\text{raw}} = \{\textbf{X}_1^{\text{raw}}, \dots, \textbf{X}_m^{\text{raw}}\} \), where each \( \textbf{X}_i^{\text{raw}} \) denotes a univariate sequence. $k$ denotes the $k$-tuple length and the \( \mathcal{F} = \{\mathrm{DISC}_1, \dots, \mathrm{DISC}_m\} \) represents the discretization functions for each dimension which can be shared or unique.
First, an empty graph is initialized \( G = (\mathcal{V}, \mathcal{E}) \) (line 2). For each input dimension \( i \), the data in that dimension is discretized \( \mathrm{DISC}_i \) (line 3-6). A sliding window of size \( k \) is applied to the discretized data to generate $k$-tuples \( R_i^t \) and \( K_i^t \) (lines 9-10). Extracted $k$-tuple is used to generate a prefix \( u_i \) and a suffix node \( v_i \) and its feature set \( F_{u_i} \) or \( F_{v_i} \) is initialized (lines 11–20). These feature sets are always updated to track which raw sequences contributed to each node (lines 21–22). 
If the edge \( (u_i, v_i) \) does not exist in \( \mathcal{E} \), then it is initialized with the edge weight \( w = 1 \), otherwise $w$ is incremented each time it is visited to reflect the frequency (lines 23–28). At the initial time step $t=0$, all prefix nodes \( u_i \) across dimensions are connected using a bidirectional hyper-tuple edge to capture the inter-dimensional dependencies (lines 30–33). For all subsequent timesteps, hyper-tuple edges are formed between suffix nodes \( v_i \) as well (lines 35–37). Finally, the constructed MdBG (\( G \))  is returned (line 39).

\begin{table}[h!]
\centering
\caption{Dataset Partition Sizes}
\label{tab:partition}
\begin{tabular}{@{}lcccc@{}}
\toprule
\textbf{Dataset} & \textbf{Train} ($S$) & \textbf{Validation} & \textbf{Test}  & \textbf{Dimensions} ($D$) \\
\midrule
ETTh1 \& ETTh2 & 8,533 & 2,785 & 2,785 & 7 \\
ETTm1 \& ETTm2 & 34,453 & 11,425 & 11,425 & 7 \\
\bottomrule
\end{tabular}
\end{table}

\begin{table}[h!]
\centering
\caption{MdBG sizes constructed from training data for each $\alpha$ (20, 25, 30). }
\label{tab:graph_sizes}
\begin{tabular}{@{}lccc@{}}
\toprule
\textbf{Dataset} & \textbf{Nodes} & \textbf{Edges} & \textbf{$\alpha$} \\
\midrule
ETTh1 & 4,137 / 6,044 / 8,104 & 207,445 / 263,494 / 304,185 & 20 / 25 / 30 \\
ETTm1 & 3,835 / 5,678 / 7,918 & 306,653 / 454,542 / 604,859 & 20 / 25 / 30 \\
ETTm2 & 1,502 / 2,215 / 3,044 & 102,003 / 162,246 / 241,866 & 20 / 25 / 30 \\
ETTh2 & 1,708 / 2,520 / 3,540 & 93,531 / 140,562 / 180,012 & 20 / 25 / 30 \\
\bottomrule
\end{tabular}
\end{table}

\begin{algorithm}[!ht]
\caption{MdBG Construction Algorithm}
\label{alg:mdBG-construction}
\DontPrintSemicolon
\SetAlgoLined
\setcounter{AlgoLine}{0}

\Input{
$\mathcal{D}^{\text{raw}} = \{\textbf{X}_1^{\text{raw}}, \textbf{X}_2^{\text{raw}}, \dots, \textbf{X}_m^{\text{raw}}\}$: Aligned multivariate time series,\\
$k \in \mathbb{N}$: Desired $k$-tuple length,\\
$\mathbf{F} = \{\textsc{DISC}_1, \dots, \textsc{DISC}_m\}$: Discretizers per dimension
}

\Output{
$G = (\mathcal{V}, \mathcal{E})$: Constructed Multivariate de Bruijn Graph
}

\SetKwFunction{FMain}{MdBGConstruction}

\Fn{\FMain{$\mathcal{D}^{\text{raw}}, k, \mathbf{F}$}}{
    Initialize empty graph: $G \gets (\mathcal{V} = \varnothing, \mathcal{E} = \varnothing)$\;
    
    \For{$i \gets 1$ \KwTo $m$}{
        $\textsc{DISC}_i \gets \mathbf{F}[i]$
        
        $\textbf{X}_i^{\text{disc}} \gets \textsc{DISC}_i(\textbf{X}^{\text{raw}}_i)$\;
    }

    \For{$t \gets 0$ \KwTo $T - k$}{
        \For{$i \gets 1$ \KwTo $m$}{
            $K_i^t \gets \textbf{X}_i^{\text{disc}}[t : t + k]$\;
            
            $R_i^t \gets \textbf{X}_i^{\text{raw}}[t : t + k]$\;
            
                $u_i \gets (i, K_i^t[0 : k{-}1])$\;
                
                $v_i \gets (i, K_i^t[1 : k])$\;

                \If{$u_i \notin \mathcal{V}$}{
                    Add $u_i$ to $\mathcal{V}$\;
                    
                     $ \mathcal{F}_{u_i} = \varnothing$
                }
                \If{$v_i \notin \mathcal{V}$}{
                    Add $v_i$ to $\mathcal{V}$\;

                     $ \mathcal{F}_{v_i} = \varnothing$

                }
                
                $ \mathcal{F}_{u_i} = \mathcal{F}_{u_i} \cup \{ R_i^t[0 : k{-}1]\}$
                    
                $ \mathcal{F}_{v_i} =  \mathcal{F}_{v_i} \cup  \{ R_i^t[1 : k]\}$

                \If{$(u_i, v_i) \notin \mathcal{E}$}{
                    Add directed edge: $u_i \xrightarrow{w = 1} v_i$\;
                }
                \Else{
                    Increment edge weight: $w \gets w + 1$\;
                }
            
        }

        \If{$t = 0$}{
            \ForAll{distinct pairs $(u_i, u_j)$ from prefix nodes}{
                Add bidirectional hyperedge: $u_i \leftrightarrow u_j$ with type \texttt{hyper}\;
            }
        }

        \ForAll{distinct pairs $(v_i, v_j)$ from suffix nodes}{
            Add bidirectional hyperedge: $v_i \leftrightarrow v_j$ with type \texttt{hyper}\;
        }
    }

    \Return $G$\;
}
\end{algorithm}

\clearpage

\section{Experimental Details}
\subsection{Evaluation Metrics}
To evaluate the performance of the constructed model, two standard regression metrics are used: mean squared error (MSE) and mean absolute error (MAE).

\textit{MSE} is defined as:
\[
\mathrm{MSE} = \frac{1}{n} \sum_{i=1}^{n} (y_i - \hat{y}_i)^2
\]

\textit{MAE} is defined as:
\[
\mathrm{MAE} = \frac{1}{n} \sum_{i=1}^{n} |y_i - \hat{y}_i|
\]
where \( y_i \) and \( \hat{y}_i \) denote the ground-truth and the predicted value, and \( n \) represents the total number of predictions.
\subsection{Extended Experimental Results}
In Table \ref{tab:full_results}, MSE and MAE of state-of-the-art (SoTA) forecasting models across on four benchmark datasets with the prediction length $PL \in \{96, 192, 336, 720\}$ are presented. The DRAGON model architecture has shown improved performance over the original TimesNet, in most configuration settings. 

All experiments use an input sequence length of 12. We employ three DRAGON modules with order $k = 4$ and discretization alphabet sizes $\alpha = 20$, $25$, and $30$, respectively for all layers/dimensions. Similarly, a uniform discretization function is used, which assigns equal-width bins across all features and dimensions for every layer. Each DRAGON module incorporates two layers of graph attention, and Graph Diffusion Convolution (GDC) is applied with a top-$k$ value of 32. For the downstream TimesNet model, the ``optimal'' hyperparameters recommended for each benchmark dataset and forecasting horizon is adopted. The graph embedding dimension $G$ is consistently set equal to the model channel size $C$. Our experimental pipeline, along with the hyperparameter configurations for all models, follows the TSLib framework~\cite{wu2022timesnet}, available at: \url{https://github.com/thuml/Time-Series-Library}.

The partition sizes for training, validation, and test splits across all datasets are summarized in Table \ref{tab:partition}. The resulting MdBG graph sizes (in terms of number of nodes and edges) for each alphabet size $\alpha$ are reported in Table \ref{tab:graph_sizes}.

The graph construction algorithm is implemented using the NetworkX library \cite{SciPyProceedings_11}, and the resulting graphs are subsequently converted to data objects compatible with PyTorch Geometric \cite{fey2019fastgraphrepresentationlearning}. All experiments are conducted on a single NVIDIA A100 GPU with 42 GB of memory.

\begin{table*}
    \centering
        \caption{MSE and MAE values of different models across different forecast horizons tested on four datasets. Best-performing values are highlighted in bold, and the second best is underlined.}
        \label{tab:full_results}
        \scalebox{0.9}{
    \begin{tabular}{ll|ll|ll|ll|ll}
        \toprule
        ~ & ~ & \multicolumn{2}{|c|}{\textbf{ETTh1}}  
        & \multicolumn{2}{|c|}{\textbf{ETTh2}}  
        & \multicolumn{2}{|c|}{\textbf{ETTm1}}  
        & \multicolumn{2}{|c}{\textbf{ETTm2}} \\
        \textbf{model} & \textbf{PL} & \textbf{MSE} & \textbf{MAE} & \textbf{MSE} & \textbf{MAE} & \textbf{MSE} & \textbf{MAE} & \textbf{MSE} & \textbf{MAE} \\ 
        \midrule
        Transformer & \multirow{2}{*}{\textbf{96}}  & 0.775 & 0.667 & 1.147 & 0.801 & 0.762 & 0.584 & 0.484 & 0.535 \\ 
        TSMixer & & 0.867 & 0.678 & 1.013 & 0.823 & 0.805 & 0.61 & 0.33 & 0.438 \\ 
        Crossformer & & 0.594 & 0.554 & 0.81 & 0.63 & \textbf{0.618} & 0.516 & 0.335 & 0.404 \\ 
        FiLM & & 0.669 & 0.521 & 0.349 & 0.37 & 0.971 & 0.591 & 0.234 & 0.309 \\ 
        Nonstationary & & 0.589 & 0.502 & 0.361 & 0.384 & 0.682 & 0.513 & 0.219 & 0.291 \\ 
        PatchTST & & 0.589 & 0.489 & 0.345 & 0.369 & 0.946 & 0.587 & 0.231 & 0.306 \\ 
        TimeMixer & & 0.583 & 0.484 & \uline{0.34} & 0.367 & 0.886 & 0.568 & 0.225 & 0.301 \\ 
        Autoformer & & 0.635 & 0.531 & 0.353 & 0.383 & 0.778 & 0.565 & 0.212 & 0.304 \\ 
        TimeXer & & \textbf{0.543} & \textbf{0.470} & \textbf{0.337} & \textbf{0.362} & 0.833 & 0.551 & 0.22 & 0.295 \\ 
        TimesNet & & 0.59 & 0.495 & 0.345 & 0.366 & 0.679 & \uline{0.506} & \uline{0.201} & \uline{0.276} \\ 
        DRAGON (Ours) & & \uline{0.552} & \uline{0.479} & 0.34 & \uline{0.363} & \uline{0.664} & \textbf{0.499} & \textbf{0.2} & \textbf{0.275} \\ 
        \midrule
        Transformer & \multirow{2}{*}{\textbf{192}} & 0.819 & 0.675 & 3.095 & 1.312 & 0.772 & 0.623 & 1.052 & 0.77 \\ 
        TSMixer & & 0.975 & 0.741 & 2.419 & 1.303 & 0.87 & 0.66 & 0.49 & 0.549 \\ 
        Crossformer & & 0.631 & 0.578 & 1.667 & 0.899 & \textbf{0.693} & 0.576 & 0.438 & 0.474 \\ 
        FiLM & & 0.711 & 0.543 & 0.446 & 0.422 & 1.015 & 0.616 & 0.305 & 0.352 \\ 
        Nonstationary & & 0.665 & 0.547 & 0.504 & 0.469 & 0.752 & 0.548 & 0.292 & \uline{0.33} \\ 
        PatchTST & & 0.637 & 0.516 & 0.442 & 0.421 & 0.995 & 0.613 & 0.303 & 0.35 \\ 
        TimeMixer & & 0.621 & 0.509 & \uline{0.438} & 0.42 & 0.925 & 0.591 & 0.297 & 0.346 \\ 
        Autoformer & & 0.655 & 0.55 & 0.451 & 0.438 & 0.807 & 0.582 & \uline{0.288} & 0.35 \\ 
        TimeXer & & \textbf{0.593} & \textbf{0.499} & \textbf{0.437} & \textbf{0.417} & 0.899 & 0.582 & 0.292 & 0.338 \\ 
        TimesNet & & 0.626 & 0.514 & 0.449 & 0.425 & 0.727 & \uline{0.533} & 0.289 & 0.337 \\ 
        DRAGON (Ours) & & \uline{0.601} & \uline{0.505} & 0.44 & \uline{0.417} & \uline{0.701} & \textbf{0.523} & \textbf{0.274} & \textbf{0.321} \\ 
        \midrule
        Transformer & \multirow{2}{*}{\textbf{336}} & 1.048 & 0.804 & 2.854 & 1.335 & 0.911 & 0.699 & 1.166 & 0.816 \\ 
        TSMixer & & 1.067 & 0.792 & 2.765 & 1.403 & 0.957 & 0.721 & 0.784 & 0.717 \\ 
        Crossformer & & 0.737 & 0.627 & 2.837 & 1.364 & \textbf{0.774} & 0.628 & 0.889 & 0.688 \\ 
        FiLM & & 0.74 & 0.563 & 0.502 & 0.464 & 1.054 & 0.638 & 0.377 & 0.392 \\ 
        Nonstationary & & 0.808 & 0.621 & 0.749 & 0.593 & 0.799 & \uline{0.572} & 0.481 & 0.434 \\ 
        PatchTST & & 0.67 & 0.538 & 0.501 & \uline{0.462} & 1.047 & 0.639 & 0.374 & 0.39 \\ 
        TimeMixer & & 0.659 & 0.532 & \textbf{0.498} & 0.465 & 0.963 & 0.613 & 0.372 & 0.389 \\ 
        Autoformer & & 0.657 & 0.564 & 0.515 & 0.483 & 0.9 & 0.627 & 0.366 & 0.395 \\ 
        TimeXer & & \uline{0.646} & 0.527 & \uline{0.498} & \textbf{0.459} & 0.919 & 0.599 & 0.363 & 0.379 \\ 
        TimesNet & & 0.657 & 0.533 & 0.512 & 0.472 & 0.892 & 0.599 & \textbf{0.349} & \textbf{0.366} \\ 
        DRAGON (Ours) & & \textbf{0.636} & \textbf{0.525} & 0.503 & 0.464 & \uline{0.786} & \textbf{0.564} & \uline{0.35} & \uline{0.369} \\ 
        \midrule
        Transformer & \multirow{2}{*}{\textbf{720}} & 0.995 & 0.786 & 3.062 & 1.429 & 1.11 & 0.782 & 2.914 & 1.256 \\ 
        TSMixer & & 1.1 & 0.813 & 2.849 & 1.414 & 1.03 & 0.768 & 2.023 & 1.188 \\ 
        Crossformer & & 0.801 & 0.678 & 2.614 & 1.345 & \textbf{0.785} & 0.633 & 2.357 & 1.133 \\ 
        FiLM & & 0.729 & 0.575 & 0.504 & 0.476 & 1.083 & 0.659 & 0.483 & 0.447 \\ 
        Nonstationary & & 0.905 & 0.675 & 0.69 & 0.578 & 0.822 & \uline{0.595} & 0.68 & 0.54 \\ 
        PatchTST & & 0.67 & 0.556 & \uline{0.502} & \uline{0.475} & 1.069 & 0.658 & 0.481 & 0.446 \\ 
        TimeMixer & & \uline{0.668} & \uline{0.55} & 0.513 & 0.483 & 0.993 & 0.633 & 0.477 & 0.445 \\ 
        Autoformer & & 0.676 & 0.58 & 0.515 & 0.494 & 0.932 & 0.638 & 0.476 & 0.455 \\ 
        TimeXer & & 0.67 & 0.552 & 0.506 & 0.476 & 0.967 & 0.624 & \uline{0.471} & \uline{0.437} \\ 
        TimesNet & & 0.681 & 0.559 & 0.523 & 0.489 & 1.1 & 0.675 & 0.488 & 0.451 \\ 
        DRAGON (Ours) & & \textbf{0.627} & \textbf{0.533} & \textbf{0.5} & \textbf{0.474} & \uline{0.819} & \textbf{0.583} & \textbf{0.455} & \textbf{0.428} \\ 
        \bottomrule
    \end{tabular}
    }
\end{table*}
\subsection{Future Work}

We plan to extend the DRAGON framework in several promising directions. The first step is to improve its scalability in terms of graph construction and masking strategies to better accommodate long-term forecasting and high dimensional time series data. Subsequently, we aim to conduct more extensive experiments across diverse benchmark datasets from various domains and multiple downstream models to assess generalization. Beyond forecasting, we intend to transform DRAGON into a multimodal architecture by incorporating support for additional downstream tasks, including  anomaly detection, classification, and imputation. Ultimately, our long-term goal is to evolve DRAGON into a foundation model by training it on large-scale, heterogeneous time series data in a self-supervised manner.


\end{document}